\documentclass{article}
\usepackage{spconf,amsmath,graphicx,hyperref}
 \usepackage{float}
\usepackage{cite}
\usepackage{algorithmic}
\usepackage{textcomp}
\usepackage{xcolor}
\usepackage{booktabs}
\usepackage{pifont}
\usepackage{url}
\usepackage{multirow}
\usepackage[numbers]{natbib}
\usepackage[table]{xcolor}

\title{Cross-Corpus and Cross-domain Handwriting Assessment of NeuroDegenerative Diseases via Time-Series-to-Image Conversion}
\name{Gabrielle Chavez$\ddagger$, Laureano Moro-Velazquez$\ddagger$, Ankur Butala$\star \dagger$, Najim Dehak$\ddagger$, Thomas Thebaud$\ddagger$}
\address{$\ddagger$Department of Electrical and Computer Engineering, The Johns Hopkins University, Baltimore MD, USA\\
$\star$Department of Psychiatry and Behavioral Sciences, The Johns Hopkins University, Baltimore MD, USA\\
$\dagger$Department of Neurology, The Johns Hopkins University, Baltimore MD, USA}
\begin{document}
%
\maketitle
\begin{abstract}
Handwriting is significantly affected by neurological disorders (ND) such as Parkinson's disease (PD) and Alzheimer's disease (AD). Prior works have analyzed handwriting tasks using feature-based approaches or computer-vision techniques, but these methods have struggled to generalize across multiple datasets, particularly between temporal features represented as time-series and images.
We propose a framework that leverages both time-series and images of handwriting through a joint classifier, based on a ResNet50 pretrained on ImageNet-1k. 
Binary classification experiments demonstrate  state-of-the-art performances on existing time-series and image datasets, with significant improvement on specific drawing and writing tasks from the NeuroLogical Signals (NLS) dataset.  In particular, the proposed model demonstrates improved performance on \textit{Draw Clock} and \textit{Spiral} tasks. Additionally, cross-dataset and multi-dataset experiments were consistently able to achieve high F1 scores, up to 98 for PD detection, highlighting the potential of the proposed model to generalize over different forms of handwriting signals, and enhance the detection of motor deficits in ND.

\end{abstract}
\section{Introduction}
\label{sec:intro}
Neurodegenerative diseseases (ND) like Parkinson’s disease (PD) and Alzheimer's Disease (AD) affect both motor and cognitive function \cite{vessio2019dynamic}. 
PD is characterized by motor and non-motor impairments that disrupt daily life. 
Among the motor symptoms, handwriting is particularly sensitive to subtle changes in motor control, including tremor, bradykinesia, and micrographia, making it a valuable marker of disease progression \cite{thomas2017handwriting}. 
In contrast, AD primarily affects cognitive function, leading to memory loss and difficulty in spatial reasoning \cite{fernandes2023handwriting}. 
These impairments also manifest in handwriting through poor spatial organization and diminished motor control \cite{chen2024cognitive}. 
One widely used handwriting task is the \textit{spiral} drawing task, which has long been used as a simple clinical tool to capture bradykinesia and tremor\cite{aversano2022early} \cite{farhah2024utilizing}. 
With modern digitized tablets, it is now possible to record high-resolution temporal features capturing the kinematics, pressure, and dynamics of handwriting movements. 

Previous studies \cite{9207087, Drot_r_2016, thebaud2024explainable, laouedj2025detecting} have approached handwriting analysis using either image-based methods, which capture spatial irregularities from static samples, or time-series methods, which extract dynamic features such as velocity, acceleration, and pressure. 
Image-based approaches have shown strong performance; for example, Razzak et al. \cite{9207087} achieved 98\% accuracy on HandPD, NewHandPD, and Parkinson’s Drawing datasets using a ResNet50 CNN–SVM framework on all the available tasks. 
However, such methods often generalize poorly to time-series datasets, achieving only 37.5\% accuracy on PaHaW \cite{9207087}. 
Time-series methods, such as those of Gazda et al. \cite{gazda2021multiple}, have improved performance by incorporating mediator datasets and multiple fine-tuned CNNs, but at the cost of additional data requirements and model complexity. 

In this article, we propose a new framework with a simple and efficient solution: transforming handwriting time-series into image representations and classifying them using the same pretrained ResNet50 feature extractor followed by a multilayer perceptron (MLP), thereby unifying spatial and temporal information within a single joint model.

The contributions of this article are the joint modeling of spatial and temporal features using a unique classifier,
achieving state-of-the-art performance on the NLS dataset across multiple tasks and demonstrating cross-dataset and multi-dataset generalization.

\section{Materials and Methods}

\subsection{Datasets}
In this study, we used five datasets: three image-based datasets (\textit{HandPD, NewHandPD, and Parkinson's Drawing}), two time-series handwriting dataset (\textit{PaHaW and NLS}). 

The \textbf{HandPD} dataset was collected at Sao Paulo State University at the Faculty of Medicine of Botucatu \cite{pereira2015step}. 
It consists of 92 participants, 74 with PD and 18 control subjects (CTL). Each individual was asked to trace a spiral, meander, or draw circles. 
The \textbf{NewHandPD} was collected in the same manner as the HandPD dataset, but using a pen with embedded sensors to also capture its tilt and acceleration, and includes 37 people with PD and 35 CTL \cite{PereiraSIBGRAPI:16}. Both the spiral template and spiral tracing were visible in the photo for these datasets. 

The \textbf{Parkinson's Drawing} dataset (\textit{ParkD})~\cite{zham2017distinguishing} recorded dynamic handwriting from 27 people with PD and 28 CTL drawing spirals via a digital tablet with a paper placed on top. The participants traced an archimedean spiral, but the template was not visible in the photo. 

The \textbf{NeuroLogical Signals} (\textit{NLS}) dataset was collected at Johns Hopkins Hospital and includes samples from synchronized speech, eye-tracking, and handwriting tasks, from participants diagnosed with various degenerative diseases. 
In this study, we focus on the handwriting component which consist 13 tasks performed by 35 PD, 42 CTL, and 21 AD \cite{thebaud2024explainable}. In this study, we ignore the point tasks (maintaining a pen without touching the tablet), and keep only 10 drawing and writing tasks, which will be easier to represent as images.
Those tasks include three free-hand spiral drawing tasks, using the dominant (\textit{SpiralDom}) and non dominant hand (SpiralNonDom), and while talking (\textit{SpiralPaTaKa}), drawing a cube (\textit{CopyCube}), an image (\textit{CopyImage}), or a clock (\textit{DrawClock}), copying a text (\textit{CopyText}), while reading it (\textit{CopyReadText}), writing freely (\textit{FreeWrite}) and solving algebra equations (\textit{Numbers}).
The \textbf{PaHaW} dataset was collected using a digital tablet and a spiral template, however only recorded the drawing. The dataset is composed of spirals from 37 PD and 38 CTL \cite{drotar2016evaluation}.

\begin{figure}[ht]
  \centering
  \includegraphics[width=1\columnwidth]{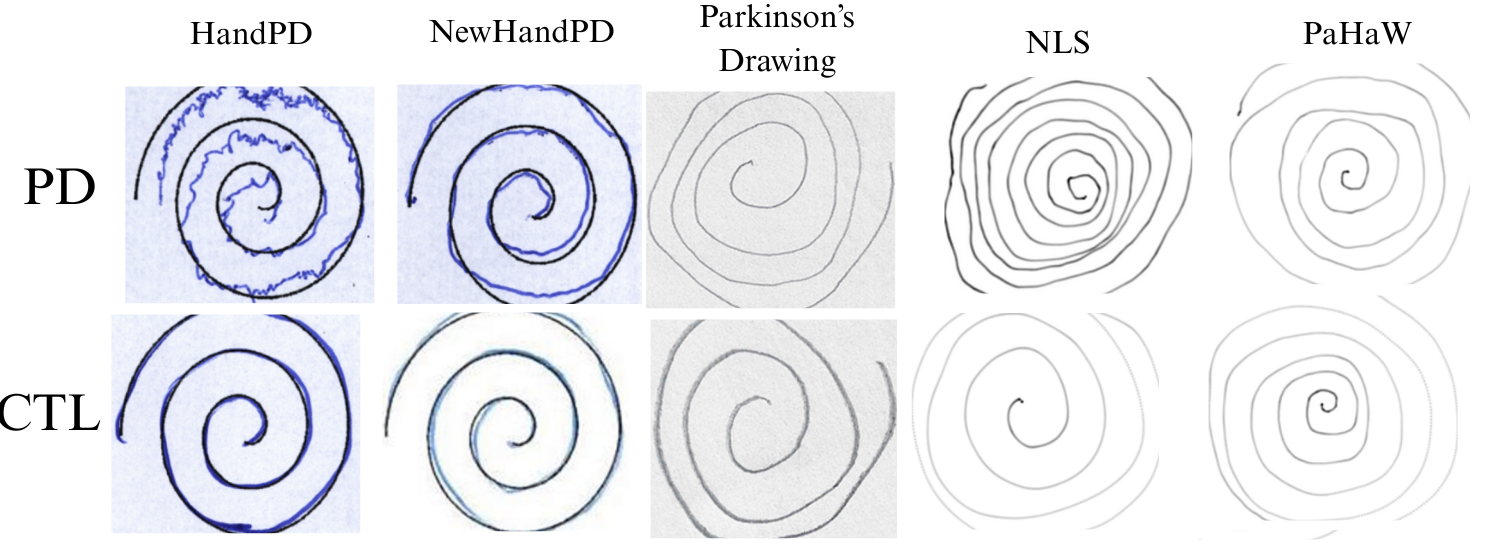}
    \vspace{-5mm}
  \caption{Samples of spirals from each dataset after pre-processing. \textit{NLS} and \textit{PaHaW} show the images created from the times series.}
  \label{fig:Spirals}
    \vspace{-5mm}
\end{figure}

The total number of samples per category and per dataset is available in Table \ref{tab:data}. Figure \ref{fig:Spirals} includes some examples of spirals obtained from the five datasets.

\begin{table}[ht]
  \centering
    \vspace{-5mm}
  \caption{Distribution of the different datasets at hand. The $\star$ marker indicates when the dataset included a template in the photo.}
  \label{tab:data}
  \resizebox{\linewidth}{!}{
  \begin{tabular}{l c ccc ccc}
  \toprule
  Dataset & Type & \multicolumn{3}{c}{\# Participants} & \multicolumn{3}{c}{\# Samples} \\
   &  & CTL & PD & AD &  CTL & PD & AD \\
  \midrule
   ParkD\cite{zham2017distinguishing} & \multirow{3}{*}{Images} & 28 & 27 & - & 51 & 51 & - \\
   $\star$HandPD\cite{pereira2015step} & &  18 & 74 & - & 72 & 296 & - \\
   $\star$NewHandPD\cite{PereiraSIBGRAPI:16} & & 35 & 31 & - & 140 & 124 & -\\
  \midrule
   PaHaW\cite{drotar2016evaluation} & \multirow{2}{*}{Time} & 38 & 37 & - & 302 & 295 & -\\
   NLS - All tasks \cite{thebaud2024explainable} & \multirow{2}{*}{Series} & 42 & 35 & 21 & 712 & 526 & 124  \\
   NLS - Spirals \cite{thebaud2024explainable} & & 40 & 33 & 20 & 251  & 201 & 54  \\
  \bottomrule
  \end{tabular}
  }
    \vspace{-5mm}

\end{table}

\subsection{Preprocessing}
Each dataset has been preprocessed according to its modality. 
For image-based datasets, all images, originally a 1:1 ratio, were linearly compressed to $224\times224$ pixels, scaled the luminosity to 275, and lastly applied a Gaussian blur filter with a radius of 1 to reduce noise and highlight structural patterns.

For time-series datasets, we used the available pressure (P) and the coordinates to construct an 1:1 ratio, $224\times224$ image representation using the Matplotlib package\footnote{\url{https://matplotlib.org/}}. 
Each time-series point is visualized as a black dot, positioned by its coordinates, scaled by normalized pressure, and rendered at 10\% opacity to reflect drawing speed. Rapid movements produce faint trajectories, whereas slower strokes yield darker patterns.



\subsection{Models}
\begin{figure} [ht]
  \centering
  \includegraphics[width=1\columnwidth]{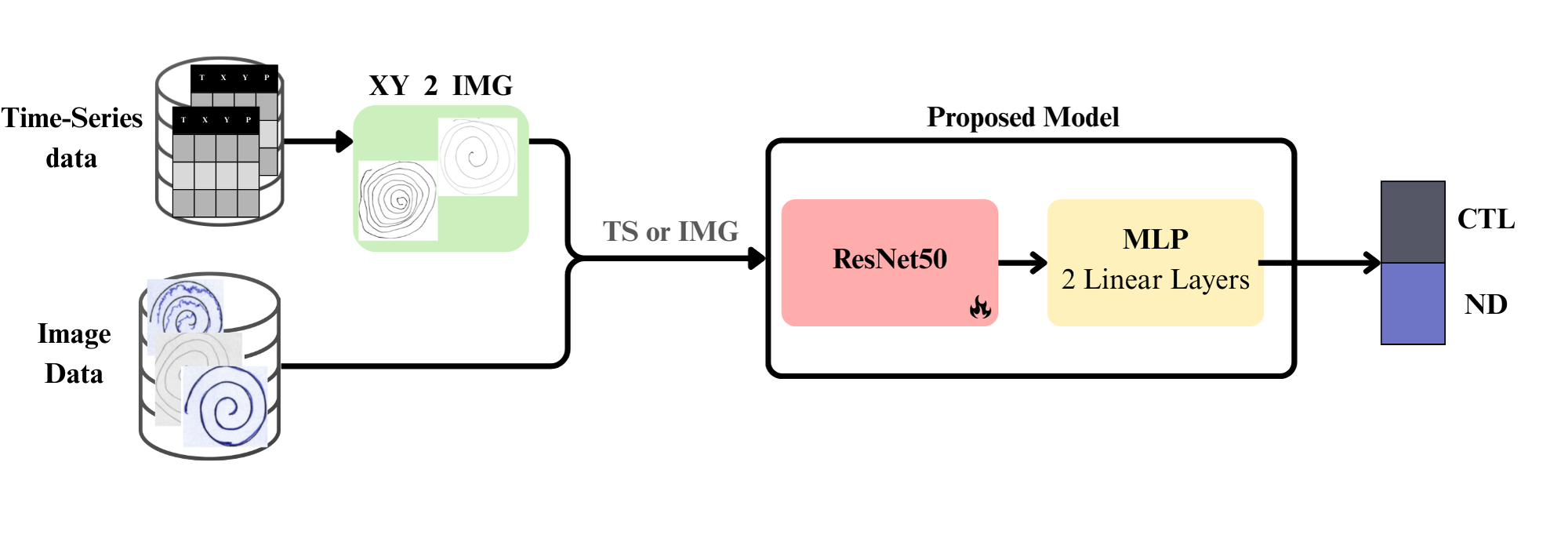}
  \vspace{-5mm}
  \caption{Block diagram of the proposed binary classification framework. Data was processed according to its modality.}
  \label{fig:archi}
  \vspace{-5mm}
\end{figure}

Our proposed model leverages a ResNet-50 v1.5\cite{he2016deep} pretrained on the ImageNet-1k~\cite{imagenet15russakovsky} dataset, with over 1.2 million of $224\times 224$ images from a thousand classes.
The model is appended with two linear layers to perform binary classification, either between AD and CTL classes or PD and CTL, and trained end to end. 
The full pipeline is shown in Figure \ref{fig:archi}.



\subsection{Metrics}
The task at hand is a binary classification with unbalanced classes, so we will use macro F-1 score ($F1$) from the predicted classes. For comparison with previous works, we will also add Unweighted Accuracy ($Acc$). All metrics are computed using the sci-kit learn package\footnote{\url{https://scikit-learn.org/stable/api/sklearn.metrics.html}}.

\section{Experiments}

\subsection{Training}
The training is performed using a 5-fold cross-validation (CV), with one fold for testing and 4 folds for training and validation.
Each fold contains the same class balance, and no subject can be part of 2 different folds.
In each fold, the dataset is partitioned into 80\%/20\% training and validation. 
Models are optimized using the AdamW optimizer~\cite{loshchilov2017decoupled} and a CrossEntropy loss, with a batch size of 16 for up to 50 epochs, using a learning rate of $10^{-5}$ for the classification head and ResNet finetuning. 
To mitigate overfitting, early stopping with a patience of 10 epochs was applied, based on validation loss. 
The metrics reported are averages across 5 folds.
The full code to reproduce the experiment will be released on github\footnote{\url{https://github.com/GabrielleChavez/ROSS-Corpus-Handwriting-NDD-TS2Img}}.

\subsection{Validation against previous works}
We first test the performances of our proposed model on the image datasets (HandPD, NewHandPD and ParKD), and compare with multiple existing baselines to verify the effectiveness of our classifier on images.

\subsection{Application to NLS}
The second series of experiment explores the effectiveness of our time-series to image conversion on the individual tasks of the NLS dataset.
A total of 20 experiments is conducted across the 10 NLS tasks, for the binary classification of PD against CTL, and AD against CTL. 
Each task and each comparison is trained and tested independently to evaluate the effectiveness of the proposed technique.
We compare the results obtained by our technique to previous results on the same dataset, using either explainable features~\cite{thebaud2024explainable} or a BiLSTM network~\cite{laouedj2025detecting} directly on the time-series.
 
\subsection{Cross-dataset and multi-datasets experiments}
To evaluate the generalization of the proposed method, we perform a series of experiments, training on 1 or more dataset, and evaluating separately on each datasets.
Three types of experiments are performed: \textbf{mono-dataset} experiments train on one dataset and evaluate on all datasets, \textbf{cross-datasets} experiments train on all datasets but one, and evaluate on the one missing, and \textbf{multi-dataset} experiments train on all datasets and test on all datasets.
The \textbf{multi-dataset-spiral} experiment only uses the spiral subset of NLS, while the \textbf{multi-dataset-all} uses the complete NLS dataset.
The mono and multi-datasets experiments are performed using a 5-folds cross-validation across all datasets, each model is trained only the training folds of the selected sets, then tested on the testing fold of each dataset, and finally the metrics are averaged for each dataset across the 5 folds.
The cross-dataset experiment is trained once on the 5 training sets and evaluated on the unused dataset.



\section{Results and Discussion}
In this section, we present the results of classifying ND versus CTL handwriting samples using a ResNet50 + MLP architecture, evaluate the effectiveness of converting time-series data into image representations for classification, and application between different datasets.

\subsection{Image classification results}
Table \ref{tab:previous_work} compares the performance of our method against baseline results across three image-based PD datasets. 
The proposed approach shows comparable results with the state-of-the-art approaches on each dataset, confirming the efficiency of the proposed method.

\begin{table}[H]
  \centering
    \vspace{-5mm}
  \caption{Comparison of our method with related works on three image datasets of spirals. *\textit{ParkD computed an handcrafted index from explainable metrics which was used as a score to discriminate between PD and CTL}}
  \begin{tabular}{l l c c }
    \toprule
    \textbf{Dataset} & \textbf{Model} &\textbf{$Acc$}\\
    \midrule
    & Na\"{\i}ve Bayes \cite{pereira2015step} & 78.9  \\
    HandPD\cite{pereira2015step} & ResNet50\cite{9207087} &  81.1\\
    &ResNet50+MLP(ours) & 88.0 \\
    \midrule
    & ImageNet \cite{PereiraSIBGRAPI:16} & 85.44 \\
    NewHandPD~\cite{PereiraSIBGRAPI:16}& ResNet50\cite{9207087} & 88.5 \\
    & ResNet50+MLP(ours) & 88.3 \\
    \midrule
    & Explainable Index*~\cite{zham2017distinguishing}& 79.1 \\
    ParkD~\cite{zham2017distinguishing}&  ResNet50\cite{9207087} & 80.0\\
    & ResNet50+MLP(ours) &77.7\\
    \bottomrule
  \end{tabular}
    \vspace{-5mm}
  \label{tab:previous_work}
\end{table}

\subsection{Application to the NLS dataset}
In this series of experiments, we compare the performances of our proposed method, leveraging a pretrained ResNet50 applied on images, to previous reported methods using either explainable features extracted from times-series~\cite{thebaud2024explainable} or using directly a BiLSTM classifier on the spectrogram of the time series~\cite{laouedj2025detecting}.
Table \ref{tab_PD_CTL_ti} shows the comparison on PD vs CTL binary classification, and Table \ref{tab_AD_CTL_ti} shows the comparison on AD vs CTL binary classification.

\subsubsection{PD vs CTL} 
While the overall improvements were not statistically significant on PD vs CTL, the results demonstrate that representing time-series data as images is a viable strategy, achieving performance comparable to established techniques. 
Notably, our framework improved accuracy on specific tasks, achieving a 10.9\% accuracy absolute gain on the \textit{Spiral Dom} task and a 14.0\% absolute gain on the \textit{Spiral NonDom} task. 

\begin{table}[H]
  \centering
  \vspace{-5mm}
  \caption{Comparison of PD vs CTL classification performances across tasks using the proposed time-series-to-image framework versus prior time-series classification method. * indicates that "Nondom" corresponds to the task labeled "Left" and "Dom" corresponds to the tasks labeled "Right".}
  \resizebox{\columnwidth}{!}{
  \begin{tabular}{l cc | c | cc}
  \toprule
    \multirow{2}{*}{\textbf{Task}} & 
    \multicolumn{2}{c|}{\textbf{Thebaud et.al \cite{thebaud2024explainable}}} & 
    \multicolumn{1}{c|}{\textbf{Laouedj et.al \cite{laouedj2025detecting}}} & 
    \multicolumn{2}{c}{\textbf{Proposed}} \\
   & $Acc$ & $F1$
   & $F1$
   & $Acc$ & $F1$ \\
    \midrule
    Spiral Dom & 54 & 52 & 52.6* & \textbf{64.9} & \textbf{58.8}\\
    Spiral NonDom & 41 & 41 & \textbf{62.3*} & \textbf{55.0} & 58.6\\
    Spiral Pataka & \textbf{60} & \textbf{60} & 52.5 & 54.7 & 49.5\\
    \midrule
    Numbers & \textbf{55} & 55 & \textbf{56.7} & 51.0 & 42.2\\
    Copy Text & \textbf{63} & \textbf{63} & 58.6 & 54.4 & 51.3\\
    Copy Read Text & 61 & 60 & \textbf{74.5} & 57.3 & 51.4\\
    Freewrite & 48 & 48 & 38.2 & \textbf{65.3} & \textbf{61.3} \\
    \midrule
    Draw Clock & 60 & \textbf{60} & 52.8 & \textbf{63.2} & 57.4\\
    Copy Cube & \textbf{64} & 63 & 40.0 & 61.3 & \textbf{52.9}\\
    Copy Image & 48 & 48 & 39.3 & \textbf{57.5} & \textbf{52.0}\\
    \bottomrule
  \end{tabular}
  }
    \vspace{-5mm}
  \label{tab_PD_CTL_ti}
\end{table}

\begin{table}[H]
    \centering
    \vspace{-5mm}
    \caption{Comparison of AD vs CTL classification performances across tasks using the proposed time-series-to-image framework versus prior time-series classification method. * indicates that "Nondom" corresponds to the task labeled "Left" and "Dom" corresponds to the tasks labeled "Right".}
    \resizebox{\columnwidth}{!}{
    \begin{tabular}{l cc | c | cc}
        \toprule
        \textbf{Task} & 
        \multicolumn{2}{c|}{\textbf{Thebaud et.al \cite{thebaud2024explainable}}} & 
        \textbf{Laouedj et.al \cite{laouedj2025detecting}} & 
        \multicolumn{2}{c}{\textbf{Proposed}} \\
         & $Acc$ & $F1$
         & $F1$
         & $Acc$ & $F1$\\
        \midrule
        Spiral Dom & 60 & 60 & 64.0* & \textbf{80.5} & \textbf{73.2}\\ 
        Spiral NonDom & 57 & 55 & 44.9* & \textbf{72.1} & \textbf{67.2}\\
        Spiral Pataka & 64 & 61 & 69.1 & \textbf{83.6} & \textbf{76.1}\\
        \midrule
        Numbers & 68 & 68 & 61.9 & \textbf{86.0} & \textbf{79.6}\\
        Copy Text & 81 & \textbf{82} & 51.2 & \textbf{81.9} & 76.3\\
        Copy Read Text & 67 & 67 & 62.7 & \textbf{86.5} & \textbf{80.3}\\
        Freewrite & 71 & 71 & 43.1 & \textbf{81.0} & \textbf{73.9}\\
        \midrule
        Draw Clock & 87 & \textbf{87} & 66.7 & \textbf{90.0} & 85.9 \\
        Copy Cube & 67 & 66 & 46.2 & \textbf{76.4} & \textbf{73.6}\\
        Copy Image & 82 & 82 & 42.9 & \textbf{85.6} & \textbf{83.0}\\
        \bottomrule
    \end{tabular}
    }
    \label{tab_AD_CTL_ti}
      \vspace{-5mm}

\end{table}

\subsubsection{AD vs CTL} 
The proposed framework demonstrates significant improvements in AD vs CTL classification compared to prior methods. 
In particular, \textit{writing, drawing,} and \textit{spiral tasks} exhibited the greatest performance gains. 
The spiral tasks showed consistent improvement, with \textit{Spiral Dom} achieving 80.5\% accuracy. 
Among writing tasks, \textit{Copy Text} and \textit{Copy Read Text} achieved accuracies greater than 85\%. 

Overall, these results demonstrate that even while losing the dynamic component of the signal, and some of the pressure information, the use of a pretrained image classifier allow for recovery of the discrimination of the signal, and even improving on previous works for certain tasks.



\subsection{Cross-datasets and multi-datasets experiments results}
Table \ref{tab:transfer} highlights strong variability in both within- and cross-dataset classification performance. Training and testing on the same dataset yields high F1 scores for HandPD (86.7) and NewHandPD (85.0), whereas datasets that excluded a spiral template yielded considerably low results. In contrast, the model show a strong generalization when trained on the Multi-dataset, achieving consistently higher performance across domains. This effect is as pronounced in the cross-copra domain, where the model is able to achieve high F1 scores on datasets unseen during training.

\begin{table}[ht]
  \centering
  \caption{F1 scores on PD vs CTL classification for transferability between datasets. Images training set includes HandPD, NewHandPD and ParkD, Times series training set includes NLS Spirals and PaHaW.}
  \resizebox{\columnwidth}{!}{
  \begin{tabular}{l| c c c c c}
  \toprule
  \textbf{Training set} & \textbf{HandPD} & \textbf{NewHandPD} & \textbf{ParkD} & \textbf{PaHaW} & \textbf{NLS}\\
  & &  & & & \textbf{Spirals} \\
  \midrule
    Hand PD   &  86.7 &  69.9 &  42.3 &  42.3 &  42.3 \\
    NewHandPD &  90.0 &  85.5 &  47.8 &  39.9 &  39.6\\
    ParkD   &  50.8 &  42.8 &  42.8 &  32.7  &  29.9 \\
    PaHaW   &  58.5 &  29.3 &  43.0 &  54.1 &  51.7 \\
    NLS Spirals &  47.8 &  36.1 &  33.4 &  37.5 &  54.9 \\
    \midrule
    Multi-dataset & \textbf{98.1} & \textbf{98.1 }& \textit{87.6} & 78.5 & 57.0 \\
    Multi-dataset-Spirals & \textit{92.4} & \textit{96.6} & 87.2 & \textbf{83.9} & \textbf{90.3} \\
    Cross-dataset & 88.4 & 93.9 & \textbf{91.5} & \textit{79.4} & \textit{85.7} \\
  \bottomrule
  \end{tabular}
  }
  \vspace{-10mm}
  \label{tab:transfer}
\end{table}


\section{Conclusions}
In this work, we introduced a model that jointly process images and time-series using a single classifier for ND assessment from handwriting.
By converting time-series data into images, we demonstrated that it is possible to capture both spatial and temporal patterns within a single model. 
On image-based datasets, our method achieved comparable performances to the state of the art, and on the time-series NLS dataset, we observed significant gains for AD assessment in tasks sensitive to spatial and memory impairments. 

Beyond individual datasets, our cross-dataset and multi-datasets experiments highlighted the transferability of the proposed approach: training on all datasets yielded up to 98.1 F1 score for HandPD and NewHandPD, and even cross-dataset training showed a significant improvement over training only on one. 
These findings suggest that simple time-series-to-image conversion not only allow strong performances thanks to the pretrained image classifier, but also hold promise for generalization across heterogeneous handwriting tasks.

Although our framework achieved competitive results, image-based classification only captures a fraction of dynamic handwriting systems, as it is ignoring the pressure variations, the speed and acceleration of the pen.

Images and Time-series based techniques might be exploiting complementary but distinct characteristics of handwriting, future works will try to bridge the gap between time-series and images by merging multiple systems.

\label{sec:refs}


\bibliographystyle{IEEEbib}
\bibliography{strings,refs}

\begin{thebibliography}{10}

\bibitem{vessio2019dynamic}
Gennaro Vessio,
\newblock ``Dynamic handwriting analysis for neurodegenerative disease assessment: a literary review,''
\newblock {\em Applied Sciences}, vol. 9, no. 21, pp. 4666, 2019.

\bibitem{thomas2017handwriting}
Mathew Thomas, Abhishek Lenka, and Pramod Kumar~Pal,
\newblock ``Handwriting analysis in parkinson's disease: current status and future directions,''
\newblock {\em Movement disorders clinical practice}, vol. 4, no. 6, pp. 806--818, 2017.

\bibitem{fernandes2023handwriting}
Carina~Pereira Fernandes, Gemma Montalvo, Michael Caligiuri, Michael Pertsinakis, and Joana Guimaraes,
\newblock ``Handwriting changes in alzheimer’s disease: A systematic review,''
\newblock {\em Journal of Alzheimer’s Disease}, vol. 96, no. 1, pp. 1--11, 2023.

\bibitem{chen2024cognitive}
Casey Chen, Thomas Thebaud, Ankur Butala, Najim Dehak, Laureano Moro-Velazquez, and Esther~S Oh,
\newblock ``Cognitive assessment through writing tasks,''
\newblock {\em Alzheimer's \& Dementia}, vol. 20, pp. e093186, 2024.

\bibitem{aversano2022early}
Lerina Aversano, Mario~Luca Bernardi, Marta Cimitile, Martina Iammarino, and Chiara Verdone,
\newblock ``Early detection of parkinson's disease using spiral test and echo state networks,''
\newblock in {\em 2022 International Joint Conference on Neural Networks (IJCNN)}. IEEE, 2022, pp. 1--8.

\bibitem{farhah2024utilizing}
Nesren Farhah,
\newblock ``Utilizing deep learning models in an intelligent spiral drawing classification system for parkinson’s disease classification,''
\newblock {\em Frontiers in Medicine}, vol. 11, pp. 1453743, 2024.

\bibitem{9207087}
Imran Razzak, Iqra Kamran, and Saeeda Naz,
\newblock ``Deep analysis of handwritten notes for early diagnosis of neurological disorders,''
\newblock in {\em 2020 International Joint Conference on Neural Networks (IJCNN)}, 2020, pp. 1--6.

\bibitem{Drot_r_2016}
Peter Drotár, Jiří Mekyska, Irena Rektorová, Lucia Masarová, Zdeněk Smékal, and Marcos Faundez-Zanuy,
\newblock ``Evaluation of handwriting kinematics and pressure for differential diagnosis of parkinson’s disease,''
\newblock {\em Artificial Intelligence in Medicine}, vol. 67, pp. 39–46, Feb. 2016.

\bibitem{thebaud2024explainable}
Thomas Thebaud, Anna Favaro, Casey Chen, Gabrielle Chavez, Laureano Moro-Velazquez, Ankur Butala, and Najim Dehak,
\newblock ``Explainable metrics for the assessment of neurodegenerative diseases through handwriting analysis,''
\newblock {\em arXiv preprint arXiv:2409.08303}, 2024.

\bibitem{laouedj2025detecting}
Sarah Laouedj, Yuzhe Wang, Jes{\'u}s Villalba, Thomas Thebaud, Laureano Moro-Vel{\'a}zquez, and Najim Dehak,
\newblock ``Detecting neurodegenerative diseases using frame-level handwriting embeddings,''
\newblock in {\em ICASSP 2025-2025 IEEE International Conference on Acoustics, Speech and Signal Processing (ICASSP)}. IEEE, 2025, pp. 1--5.

\bibitem{gazda2021multiple}
Matej Gazda, M{\'a}t{\'e} Hire{\v{s}}, and Peter Drot{\'a}r,
\newblock ``Multiple-fine-tuned convolutional neural networks for parkinson’s disease diagnosis from offline handwriting,''
\newblock {\em IEEE Transactions on Systems, Man, and Cybernetics: Systems}, vol. 52, no. 1, pp. 78--89, 2021.

\bibitem{pereira2015step}
Clayton~R Pereira, Danillo~R Pereira, Francisco~A Da~Silva, Christian Hook, Silke~AT Weber, Luis~AM Pereira, and Joao~P Papa,
\newblock ``A step towards the automated diagnosis of parkinson's disease: Analyzing handwriting movements,''
\newblock in {\em 2015 IEEE 28th international symposium on computer-based medical systems}. Ieee, 2015, pp. 171--176.

\bibitem{PereiraSIBGRAPI:16}
C.~R. Pereira, S.~A.~T. Weber, C.~Hook, G.~H. Rosa, and J.~P. Papa,
\newblock ``Deep learning-aided parkinson's disease diagnosis from handwritten dynamics,''
\newblock in {\em Proceedings of the SIBGRAPI 2016 - Conference on Graphics, Patterns and Images}, 2016, pp. 340--346.

\bibitem{zham2017distinguishing}
Poonam Zham, Dinesh~K Kumar, Peter Dabnichki, Sridhar Poosapadi~Arjunan, and Sanjay Raghav,
\newblock ``Distinguishing different stages of parkinson’s disease using composite index of speed and pen-pressure of sketching a spiral,''
\newblock {\em Frontiers in neurology}, vol. 8, pp. 268142, 2017.

\bibitem{drotar2016evaluation}
Peter Drot{\'a}r, Ji{\v{r}}{\'\i} Mekyska, Irena Rektorov{\'a}, Lucia Masarov{\'a}, Zden{\v{e}}k Sm{\'e}kal, and Marcos Faundez-Zanuy,
\newblock ``Evaluation of handwriting kinematics and pressure for differential diagnosis of parkinson's disease,''
\newblock {\em Artificial intelligence in Medicine}, vol. 67, pp. 39--46, 2016.

\bibitem{he2016deep}
Kaiming He, Xiangyu Zhang, Shaoqing Ren, and Jian Sun,
\newblock ``Deep residual learning for image recognition,''
\newblock in {\em Proceedings of the IEEE conference on computer vision and pattern recognition}, 2016, pp. 770--778.

\bibitem{imagenet15russakovsky}
Olga Russakovsky, Jia Deng, Hao Su, Jonathan Krause, Sanjeev Satheesh, Sean Ma, Zhiheng Huang, Andrej Karpathy, Aditya Khosla, Michael Bernstein, Alexander~C. Berg, and Li~Fei-Fei,
\newblock ``{ImageNet Large Scale Visual Recognition Challenge},''
\newblock {\em International Journal of Computer Vision (IJCV)}, vol. 115, no. 3, pp. 211--252, 2015.

\bibitem{loshchilov2017decoupled}
Ilya Loshchilov and Frank Hutter,
\newblock ``Decoupled weight decay regularization,''
\newblock {\em arXiv preprint arXiv:1711.05101}, 2017.

\end{thebibliography}

\end{document}